\pdfoutput=1

\documentclass[11pt]{article}
\usepackage[final]{acl}

\usepackage{times}
\usepackage{latexsym}

\usepackage[T1]{fontenc}
\usepackage[utf8]{inputenc}

\usepackage{microtype}

\usepackage{inconsolata}

\usepackage{multirow}
\usepackage{array}
\usepackage{makecell}
\usepackage{tabularx}

\newcolumntype{L}[1]{>{\raggedright\let\newline\\\arraybackslash\hspace{0pt}}m{#1}}
\newcolumntype{C}[1]{>{\centering\let\newline\\\arraybackslash\hspace{0pt}}m{#1}}
\newcolumntype{R}[1]{>{\raggedleft\let\newline\\\arraybackslash\hspace{0pt}}m{#1}}

\author{Milena Pustet \and Elisabeth Steffen \and Helena Mihaljević \\
         HTW Berlin, Germany \\\texttt{\{pustet,steffen,mihalje\}@htw-berlin.de} }

\begin{document}
\title{Detection of Conspiracy Theories Beyond Keyword Bias in German-Language Telegram Using Large Language Models}

\maketitle

\begin{abstract}
The automated detection of conspiracy theories online typically relies on supervised learning. However, creating respective training data requires expertise, time and mental resilience, given the often harmful content. Moreover, available datasets are predominantly in English and often keyword-based, introducing a token-level bias into the models. 
Our work addresses the task of detecting conspiracy theories in German Telegram messages. We compare the performance of supervised fine-tuning approaches using BERT-like models with prompt-based approaches using Llama2, GPT-3.5, and GPT-4 which require little or no additional training data. We use a dataset of $\sim\!\! 4,000$ messages collected during the COVID-19 pandemic, without the use of keyword filters.

Our findings demonstrate that both approaches can be leveraged effectively: For supervised fine-tuning, we report an F1 score of $\sim\!\! 0.8$ for the positive class, making our model comparable to recent models trained on keyword-focused English corpora.
We demonstrate our model's adaptability to intra-domain temporal shifts, achieving F1 scores of $\sim\!\! 0.7$.  Among prompting variants, the best model is GPT-4, achieving an F1 score of $\sim\!\! 0.8$ for the positive class in a zero-shot setting and equipped with a custom conspiracy theory definition. 

\end{abstract}

\noindent 

\section{Introduction}
Conspiracy theories (CTs) are not a new phenomenon, but digital communication on social networks and messenger services allows them to spread at an unprecedented speed and scale. 
This becomes particularly acute in times of crisis, such as the COVID-19 pandemic \cite{kou2017,shahsavari2020}, when individuals turn to simplistic narratives in an attempt to restore clarity and alleviate feelings of powerlessness \cite{sunstein2009,douglas2017}. The spread of CTs can hinder informed decision-making and erode public trust in institutions. Many conspiracy theories promote dehumanizing, racist, antisemitic, or otherwise objectionable worldviews, and have contributed to an increase in hate speech and hate crimes both online and offline \cite{gover2020,vergani2022}.  

Although the automated detection of related phenomena such as misinformation or fake news has made notable strides \cite{zhou_etal_fake_news_2020,ameur_etal_2023,chen_combating_2023}, conspiracy theories remain relatively underexplored. Moreover, prior research has predominantly focused on English-language data, commonly built through pre-filtering of corpora using keywords that introduce a bias towards a few particular CTs and rather explicit narratives. 
This limits the understanding regarding the efficacy of existing modeling approaches in broader thematic contexts, and the practical applicability of such models for civil society organizations that often monitor, e.g., entire communities rather than posts containing specific keywords. 
Our work addresses this gap by undertaking a comprehensive modeling attempt for automated CT detection in German-language texts. We leverage an annotated dataset from the pandemic time, randomly sampled from public Telegram channels known for disseminating conspiracy narratives \cite{steffen_etal_2023}, without relying on keyword-based filtering.

We compare text classification approaches using supervised fine-tuning with BERT-based models \cite{devlin2018}, and prompt-based classification using generative models including the closed models GPT-3.5 and GPT-4, and the open model Llama 2. 
Our first objective is to determine whether BERT-based models fine-tuned on a corpus obtained without keyword-based filtering can achieve a performance in a similar range as models trained on English keyword-based online datasets (RQ1). Next, we investigate the model's practical utilization by evaluating it in a wider range of channels and a different time frame within the same platform (RQ2). We then investigate whether prompt-based models can match or even surpass models obtained through supervised fine-tuning (RQ3), and explore the impact of different configurations on their performance, including zero-shot vs. few-shot, provided definition, and output constraints (RQ4). 

With regard to RQ1, we present the model \textit{TelConGBERT} which achieves a macro-averaged F1 score of score of 0.85 (0.79 for the positive and 0.9 for the negative class, respectively). This performance is close to that of other models trained to detect conspiracy theories in English language social media posts obtained through keyword-based filtering (with F1 scores around 0.85, see Section \ref{sec:related_work_supervised}).
When applying the model to data from later time ranges (RQ2), it shows moderate to good performance (F1 score of up to 0.72 for the positive class).
Regarding RQ3, both the supervised fine-tuning and the prompting approach achieve results in the same range, with no statistically significant difference. Nevertheless, the models' predictions disagree on 15\% of the test data.  
A notable observation regarding RQ4 is the superiority of zero-shot models over few-shot models, confirming the results reported by, e.g., \citet{chae_large_2023}. The best performing and most stable generative model is GPT-4, provided with a tailored expert definition of CTs, while the performance of GPT-3.5 and Llama 2 is less robust with regard to input configurations and output constraints.

\section{Related Work}
\label{sec:related_work}

In research, the term conspiracy theory is often used synonymously with disinformation, misinformation, rumors, or fake news \cite{mahl2022a}. While these phenomena can overlap (e.g., by using misinformation to support a conspiracy theory), CTs have distinct features: They assert a strong belief in a secret group intending to control institutions or even the world through intentionally causing complex, often unsolved events. \cite{mahl2022a,sunstein2009}. CTs offer alternative interpretations by attributing events to hidden powerful figures. They typically involve \textit{actors} such as corrupt elites pursuing malicious \textit{goals}, such as population control, through \textit{strategies} like microchip insertion via vaccinations \cite{samory2018a}. In the realm of social media and messaging services, complex narratives are often fragmented, especially when the audience is assumed to be partly informed \cite{sadler2021,ernst2017}.

\subsection{Supervised Fine-Tuning of (small) LMs}
\label{sec:related_work_supervised}
The increasing dissemination of conspiracy theories in the context of the COVID-19 pandemic has prompted computational efforts for their large scale detection and analysis. A fundamental step in such efforts is typically the creation of labeled datasets by human experts or crowd annotators.
Until recently, Twitter has been an important source of data. 
\citet{pogorelov_etal_2021} compiled a dataset of $\sim\!\! 10,000$ tweets containing keywords related to COVID-19 and 5G, and trained a binary classification model which attained an F1 score of 0.84. 
The dataset was later extended to the COCO dataset \cite{langguth_coco_2023}, which also contains labels indicating whether a tweet relates to or supports a mentioned CT. 

\citet{phillips_etal_2022} compiled $\sim\!\! 3,000$ texts based on keywords related to climate change, the COVID-19 virus, and the Epstein-Maxwell trial. 
A macro-F1 score of 0.9 was achieved, indicating that even smaller corpora can be sufficient in a restricted scenario\footnote{The corpus was created using the terms epsteincoverup, GhislaineMaxwellTrial, JeffreyEpstein, LolitaExpress, PedophileIsland, epsteinDidntKillHimself.}. 
\citet{moffittHuntingConspiracyTheories2021} collected a dataset of $\sim\!\! 8,000$ tweets by using search terms related to CTs. 
They fine-tuned a BERT model and the specialized COVID-Twitter-BERT model CT-BERT \cite{muller2020}, achieving an F1 score of 0.87 on a test set of 200 tweets.
CT-BERT and other models adapted for Twitter or COVID-19 have also been successfully used for the ‘FakeNews: Corona Virus and Conspiracies Multimedia Analysis Task’ \cite{pogorelov2021a} in the MediaEval challenge 2021, see, e.g., \cite{peskine2021,vaigh2021}. 

When interpreting the performance of these models, it is important to take into account that the underlying corpora were obtained through  
keyword-based filtering. This is a typical step in pipelines for the automated detection of CTs and related phenomena \cite{marcellino2021,memon2020,moffittHuntingConspiracyTheories2021,serrano2020}, 
usually deemed necessary to obtain a sufficient number of examples from the target class (sometimes even as high as 75\% in \citet{phillips_etal_2022}).  
As shown by the authors of the LOCO dataset\cite{miani_etal_2021}, CT related keywords such as `big pharma' or `NWO' can already serve as a well performing binary classifier of CT content for some types of content such as standalone web-documents.

However, as such filters narrow the scope to texts explicitly mentioning pre-defined signal terms, it is unclear whether similar performance is realistic for broader data cohorts.\footnote{It should be noted that keyword-based pre-filtering not necessarily results in a limited set of CTs, as this clearly depends on the set of selected keywords (cf. methods in \cite{miani_etal_2021}}).
Diverging from this paradigm, the TelCovACT dataset, which we utilize in this article, consists of $\sim\!\! 4,000$ messages randomly sampled from around 100 public German Telegram channels previously identified as frequently disseminating CTs and misinformation in the context of COVID-19 \cite{steffen_etal_2023}. It was annotated with regard to the occurrence of CTs, narrative components and stance.
The collection procedure ensured a decent proportion of relevant samples (around 36\%).
Furthermore, focusing on Telegram data enables researchers to analyze a domain with hardly any content moderation \cite{holzer2021,hoseini2021,salheiser2020,winter2021}, providing a haven for accounts `deplatformed' from major platforms due to spreading of  disinformation and hate speech \cite{curley2022a,zeitungsueddeutsche2021}. As such, we believe that it requires more attention from research.

\subsection{Zero-Shot and Few-Shot Classification}
The availability of advanced autoregressive Large Language Models (LLMs) stimulated research into their capacity to detect deceptive and harmful online content, including misinformation \cite{bang_multitask_2023,pan_fact-checking_2023,chen_combating_2023}, hate speech \cite{li_hot_2023}, toxic language \cite{wang_toxicity_2022}, antisemitism \cite{pustet_etal_decoding_report_6_2024}, or racism and sexism \cite{chiu_detecting_2022}. Such models enable text classification with prompts containing minimal (few-shot) or even no (zero-shot) in-context examples. 
Prompting, the design of textual instructions for the model, plays a vital role: These instructions may shape response formats, guide model focus, or offer additional information like definitions or in-context examples \cite{liu_pre-train_2021,white_prompt_2023}.

Initial evaluations show that these models can outperform human annotators in content moderation \cite{gilardi_chatgpt_2023} and political text classification   \cite{tornberg_chatgpt_2023}. 
When tasked with the detection of hateful, offensive, and toxic (HOT) content, \texttt{GPT-3.5-turbo} achieved F1 scores between 0.43 to 0.67 for the positive class of the respective HOT category, with an approximate accuracy of 80\% compared to crowdworkers' annotations \cite{li_hot_2023}. 
\citet{huang_is_2023} demonstrated ChatGPT's capability not only in identifying 80\% of implicit hateful tweets from the LatentHatred dataset \cite{elsherief-etal-2021-latent} but also in generating explanations of comparable quality to human annotators.
\citet{mendelsohn_dogwhistles_2023} evaluated GPT-3 and GPT-4 on the task of identifying `dog whistles', finding that performance varies greatly across different target groups.

Comparisons between fine-tuned small LMs and prompting-based experiments with LLMs yield inconclusive results, which vary depending on the task, corpus, and experimental setting   \cite{russo2023acti,bang_multitask_2023,pelrine2023reliable,pustet_etal_decoding_report_6_2024}.
Fine-tuned BERT-based models can compete or even outperform generative models, at significantly reduced costs \cite{chae_large_2023,mu_navigating_2023,yu_open_2023}. 
\citet{pelrine2023reliable} conduct extensive experiments on detecting misinformation, comparing small LMs with GPT-4 in settings similar to ours. GPT-4 achieves the highest performance (F1 score of 0.68) for binary classification when predicting a probabilistic score with a threshold optimized on a validation set. 
 
\citet{liu2024conspemollm} used a corpus created from the COCO dataset \cite{langguth_coco_2023} and an annotated subset of the LOCO dataset \cite{mompelat2022loco} to fine-tune an emotion-based LLM for five prompt-based classification tasks, comparing it to a number of baselines. The best model in the binary classification task achieved an F1 score of 0.74, while the ChatGPT baseline F1 score was 0.66. 
Several works use prompt-based zero shot classification with ChatGPT to establish baselines, reporting F1 scores around 0.40 \cite{lei2023identifying}, 0.66 \cite{liu2024conspemollm}, or a macro-averaged F1 score of 0.44 \cite{poddar2024covid19}.

Other findings point to certain limitations and inconsistencies of prompt-based approaches. These include the non-deterministic outputs of GPT-3 and Llama 2, as well as the substantial impact of minor prompt variations on the models' outputs \cite{reiss_testing_2023,mu_navigating_2023,khatun2023reliability}.
\citet{chae_large_2023} observed a decline in performance in few-shot scenarios compared to zero-shot settings.

Some studies focus on deceptive content in low-resource languages.  \citet{kuznetsova2023generative}, for example, conduct prompt-based experiments in Ukrainian, Russian, and English, albeit with a small dataset containing only five statements per language across five topics, including one CT statement each. The ACTI challenge \cite{russo2023acti} utilized an Italian-language Telegram dataset, resulting in models with F1 scores between 0.78 and 0.86. The data compilation procedure is similar to that of TelCovACT (that we utilize). However, the final dataset is smaller in size and appears to be skewed towards four CTs (data selection and annotation process are not fully clear). 
To the best of our knowledge, our work is the first to comprehensively evaluate prompt-based approaches for the automated detection of German-language conspiracy theories. 

\section{Data and Methods}

\subsection{Dataset}
\label{sec:dataset}
We employ the dataset \textit{TelCovACT} \cite{steffen_etal_2023}, in whose creation we participated and which is accessible upon request. It contains 3,663 German-language messages from public Telegram channels known for their opposition to pandemic countermeasures. The messages were posted between March 11, 2020, and December 19, 2021.
The dataset was annotated by an interdisciplinary research team with regards to three aspects: (1) the presence of a CT, indicated by a binary label, (2) narrative components of a CT, including actor, strategy, goal, and references to known CTs (e.g. \#NWO), and (3) the stance, which can be belief, authenticating, directive, rhetorical question, disbelief, neutral or uncertain. The models and experiments presented in this paper consider the binary task only. Around 36\% of the texts contain CTs, 95\% of which express belief in the communicated content. The two most frequently identified narrative components were strategy (72\%) and actor (64\%). Only 26\% of the records contained all of actor, strategy, and goal, indicating that the majority of narratives are fragmented.
For the positive class, we include only texts that express belief, and exclude texts that contain only a reference (such as a hashtag), in order to prevent the model from focusing solely on explicit signal words. Table \ref{tab:dataset} provides an overview of the dataset split for training and evaluation.

\begin{table}[h!]
\centering
\small
\begin{tabular}{p{2.3cm}||p{2.0cm}|p{2.0cm}} 
 Dataset& Negative class &Positive class\\
 \hline
 Train  (80\%)  & 1,873     &886\\
 Validation (10\%) &   241  &104\\
 Test (10\%) & 230 & 115\\
  \hline
 Total    &2,344 & 1,105\\
\end{tabular}
\caption{Training, validation and test dataset sizes.}
\label{tab:dataset}
\end{table}

\subsection{Supervised Fine-Tuning}
As a first step, we evaluated nine pre-trained BERT-based models to determine the most promising ones for the subsequent experiments. The models were selected from  Huggingface based on their suitability for German texts, relevance to the TelCovACT corpus, and popularity within the platform. 
Various combinations of model- and dataset-related hyperparameters were evaluated through Bayesian optimization. 
No German models specifically designed for pandemic-related documents or for data from  Telegram were found\footnote{The COVID-Twitter-BERT model \cite{muller2020} was exclusively trained on English language data.}. As previous studies have shown improved performance through further pre-training (retraining) on in-domain data \cite{beltagy2019,lee2019,nguyen2020}, we also applied this step to the pre-trained model that performed best in the initial experiment. Details on fine-tuning and retraining are provided in the appendix.
Additionally, we compared the performance of the best BERT-based model with a generative model, \texttt{GPT-3 davinci-002}, fine-tuned using default hyperparameters. To ensure the model adhered to the most probable answer, the temperature was set to 0.
To restrict the outputs to 0 and 1, we set a maximum of one token and adjusted the logit bias for the corresponding token IDs to 100.

\subsection{Prompt-Based Setting}
We evaluate the models GPT-3.5 (\texttt{gpt-3.5-turbo-0613}), GPT-4 (\texttt{gpt-4-0613}), and Llama 2 (\texttt{Llama2-70b-chat}).
Although Llama 2 was primarily trained on English data \cite{touvron_llama_2023_1}, it was selected due to the absence of scientifically evaluated open alternatives for German texts.
Preliminary experiments showed that Llama 2 has a basic comprehension of German and can differentiate texts related to CTs, justifying its inclusion. 

All GPT model experiments were carried out through OpenAI's API\footnote{\url{https://openai.com/}}, while Llama 2 was accessed via Replicate's API\footnote{\url{https://replicate.com/}}.

\subsubsection{Zero-Shot}
Experiments were conducted in two settings: a \textit{binary prediction task} with answer options limited to `Yes' and `No', and a \textit{probabilistic prediction task} that required predicting a probability score between 0 and 1.
We opted for this approach that  was also applied by \citet{li_hot_2023} to assess the model's confidence, as recent research indicates the ability of LLMs to articulate better-calibrated confidences using (numerically) verbalized probability scores compared to the internal conditional probabilities \cite{tian2023just}.
The experimental setup varied additionally in terms of the definition of CTs provided to the model: a) a custom definition based on the annotation guide used for the TelCovACT dataset, b) a 100-word version of Lorem Ipsum, and c) no definition.
The same prompt structure was used for GPT and Llama 2 to ensure comparability, with minor adjustments to achieve a parsable output with Llama 2. See Table \ref{tab:prompts_binary}
 and \ref{tab:prompts_proba} in the Appendix for the concrete prompts.
 
\subsubsection{Few-Shot}
\label{sec:methods_few_shot}
For this experiment, the model was provided with a set of in-context examples and corresponding labels. It was then tasked to classify a given text by returning the corresponding label (cf. Table \ref{tab:prompts_few} in the Appendix).
To evaluate robustness, we composed ten sets of 14 in-context examples, each comprising seven randomly selected instances for the positive and the negative class.
The sampling of positive examples reflected the distribution of narrative components (actor, strategy, goal) in the dataset, including two messages with one component, three messages with two, and two messages with three components.
While some studies propose that selecting in-context examples based on their semantic similarity with the target message can enhance performance \cite{liu_what_2021}, it may not be feasible in real-world situations, as it would require a substantial array of different examples, nullifying the advantage over supervised fine-tuning approaches. Therefore, we opted to use random sampling of in-context examples.
To avoid lengthening the input and due to cost considerations, we made the decision not to include a definition in this experiment.

\subsection{Comparison of models}

Relevant differences in model performances are tested for statistical significance using suitable tests, mainly the t test and McNemar's test, with significance level of 0.05 \cite{japkowicz_shah_2011}.

\section{Results}

\subsection{Supervised Fine-Tuning}
Based on the initial assessment, the pre-trained model \texttt{deepset/gbert-base} was selected as the most suitable. However, most models produced comparable results, suggesting their usefulness for the task.  
We present the fine-tuned model that achieved the best F1 score for the positive class and a possibly balanced precision and recall on the validation set during hyperparameter tuning. Table \ref{tab:results_testset} displays the model's performance on the test set, with an F1 score of 0.75 for the positive class and a macro-averaged F1 score of 0.82.\footnote{Replacing the cross-entropy loss with the self-adjusting dice loss during hyperparameter optimization resulted in a slightly higher recall for class 1. However, this came at the cost of a lower precision, and subsequently a lower F1 score.}
As expected, applying the same hyperparameter optimization to the additionally pre-trained model resulted in significantly higher scores: As Table \ref{tab:results_testset} shows, the F1 score on the positive class increases to 0.79, especially due to an improvement in precision.
Note that, in contrast to the previous experiments, several hyperparameter configurations yielded satisfactory results, indicating an overall improved suitability of the domain-adapted model.

The last column in Table \ref{tab:results_testset} presents the test set performance of the GPT-3 davinci model. Fine-tuned solely with standard hyperparameters, it achieves performance almost as good as the fine-tuned domain-adapted BERT-based model. In fact, the difference between these two models is not statistically significant. This demonstrates that achieving comparable performance with a model much larger than BERT requires significantly less effort in fine-tuning.

The retrained model that achieved the best F1 score among the fine-tuned models will be referred to as \texttt{TelConGBERT}.  

\begin{table}[h!]
\caption{Performance of the best fine-tuned models, for the base model \texttt{deepset/gbert-base}, the retrained model  \texttt{TelConGBERT}, and \texttt{GPT-3 davinci}. The highest scores for each metric are highlighted in bold.}
\centering
\small 
\begin{tabularx}{\linewidth}{ll|X|X|X}
 Metric & Class & Base & Retrained\: & GPT-3 \\
\hline
\multirow{2}{*}{Precision} 
& 0     & 0.87   & \textbf{0.88}   &   0.87\\
& 1     & 0.76  & \textbf{0.83}  &  \textbf{0.83}  \\
                           \hline
\multirow{2}{*}{Recall}    
& 0     & 0.89   & 0.92   &  \textbf{0.93}  \\
& 1     & 0.73   & \textbf{0.76}  &  0.71 \\
                           \hline
\multirow{2}{*}{F1 score}  
& 0     & 0.88   & \textbf{0.90}   &  0.89 \\
& 1     & 0.75   & \textbf{0.79}   &   0.77  \\
& macro     & 0.82   & \textbf{0.85}   &   0.83  \\
                           \hline
Accuracy  &       & 0.83 & \textbf{0.87}  &   0.86     \end{tabularx}
\label{tab:results_testset}
\end{table}

\subsubsection{Intra-Domain and Temporal Transfer}

To evaluate the robustness of \texttt{TelConGBERT}, we annotated two `transfer datasets' following the annotation scheme of the utilized dataset TelCovACT \cite{steffen_etal_2023}. 
The additional data was provided by \textit{Bundesarbeitsgemeinschaft `Gegen Hass im Netz' (BAG)}\footnote{\url{https://bag-gegen-hass.net/}}, an NGO that monitors hateful communication on Telegram in the long term, and has categorized a large number of Telegram channels based on their ideological stance.
Our sample covers channels categorized as conspiracism ('Konspirationismus') and right-wing extremism ('Rechtsextremismus').

For the first transfer dataset, we randomly selected 1,000 messages from these channels that were posted within the three months immediately following the time range of  TelCovACT (mid December 2021 to March 31, 2022). 
For the second set, we  sampled 1,000 messages posted between April 1, 2022, and July 31, 2023, thus extending the time frame to include more recent topics. To test the model with more intricate examples, we restricted to channels categorized under the subcategories 'QAnon' and 'conspiracy ideology' (Verschwörungsideologie).

It should be noted that both transfer sets were sampled from a wider range of channels than the TelCovACT dataset: Set 1 covers a total of 1,021 channels, out of which only 66 were represented in TelCovACT, while set 2 covers 450 channels, out of which only 46 overlap.

Messages that were considered too short after removing URLs and author handles were excluded.

Table \ref{tab:results_transfer} presents the performance of \texttt{TelConGBERT} on the two transfer datasets: 
For set 1, the model achieves an F1 score of 0.72 for the positive class and a macro-averaged F1 score of 0.84, which is close to its performance on the test set (see Table \ref{tab:results_testset}).  
For set 2, we report an F1 score of 0.67 for the positive class. The decrease in performance suggests challenges due to the broader temporal and topical scope of the data. However, the results demonstrate that \texttt{TelConGBERT} has moderate to good transferability, providing a positive answer to RQ2.

\begin{table}[h!]
\caption{Performance of \texttt{TelConGBERT} on data sourced from an expanded set of channels within a time frame following the training data.}
\centering
\small
\begin{tabularx}{\linewidth}{ll|X|X}
\makecell{\\Metric} & \makecell{\\Class} & \makecell{Transfer \\ dataset 1} & \makecell{Transfer \\ dataset 2}  \\
\hline
 \multirow{2}{*}{Support}
 & 0     & 672   & 589     \\
& 1     & 84 (11\%)   & 88 (13\%)  \\
 \hline
\multirow{2}{*}{Precision} & 0     & 0.98   & 0.96     \\
& 1     & 0.64  & 0.64  \\
                           \hline
\multirow{2}{*}{Recall}    & 0     & 0.94   & 0.94   \\
& 1     & 0.82   & 0.7   \\
                           \hline
\multirow{2}{*}{F1 score}  & 0  & 0.96   & 0.95\\
& 1     & 0.72   & 0.67  \\
& macro     & 0.84   & 0.81  \\
                           \hline
Accuracy  &  & 0.93 & 0.91  
\end{tabularx}
\label{tab:results_transfer}
\end{table}

\subsection{Zero-Shot Classification}
Table \ref{tab:results_zero_shot} presents the results of the zero-shot experiments. To binarize the probabilistic outputs, we computed an optimal threshold for each model on the validation set based on precision-recall-curves. With optimal thresholds of 0.8 for GPT-3.5, 0.7 for GPT-4 and 0.85 for Llama 2, respectively, the models appear to be sub-optimally calibrated.

\begin{table}[h!]
\caption{Zero-shot performance  by model, provided definition, and prediction type (binary vs. probabilistic). In the probabilistic setting, scores $\geq$ a model-specific threshold are assigned to class 1. Highest scores for each prediction setting are highlighted in bold.}
\centering
\small
\begin{tabularx}{\linewidth}{ll|XXXX}
 Model & Definition & F1\_0 & F1\_1 & macro F1 & Acc. \\
\hline
  \multicolumn{6}{p{6cm}}{Binary classification} \\ 
\hline
\multirow{3}{*}{GPT-3.5} & Custom  &  0.87  & 0.68 & 0.78 & 0.82 \\
& Lorem Ipsum  & 0.86 & 0.63 & 0.75 & 0.80\\
& None  & 0.87 & 0.72 & 0.8  & 0.83 \\
\hline
\multirow{3}{*}{GPT-4}  &  Custom  & \textbf{0.89} &  \textbf{0.79} & \textbf{0.84} & \textbf{0.86}\\
& Lorem Ipsum & --  & --  & -- & --\\
& None & 0.84 &  0.75 & 0.8 & 0.81 \\
                           \hline
\multirow{3}{*}{Llama 2} &  Custom  & 0.85 & 0.59 & 0.72 & 0.79 \\
& Lorem Ipsum & 0.81 &  0.08 & 0.44 & 0.68\\
& None & 0.87 & 0.63 & 0.75 & 0.81 \\
              \hline          
 \multicolumn{6}{p{6cm}}{Probabilistic classification} \\ 
 \hline
\multirow{3}{*}{GPT-3.5} & Custom  & 0.83 & 0.72 & 0.78 & 0.79 \\
& Lorem Ipsum  & 0.86 & 0.76 & 0.81 & 0.82 \\
& None  &  0.84 & 0.72 & 0.78 & 0.80 \\
\hline
\multirow{3}{*}{GPT-4} &  
Custom  &  \textbf{0.89} &  \textbf{0.79} & \textbf{0.84} &  \textbf{0.86} \\
& Lorem Ipsum  &  -- & -- & -- & --\\
& None &  0.84 & 0.74 & 0.79 & 0.80\\
\hline
\multirow{3}{*}{Llama 2} & Custom  & 0.56 & 0.60 & 0.58 & 0.58 \\
& Lorem Ipsum  &  0.71 & 0.61 &  0.66 & 0.67 \\
& None  & 0.64 & 0.63 & 0.64 & 0.63
\end{tabularx}
\label{tab:results_zero_shot}
\end{table}

The best performing model was GPT-4 with an F1 score of 0.79 for the positive class and a macro-averaged F1 score of 0.84. 
The model performs best when provided with the custom definition.\footnote{Due to its higher performance with a custom definition compared to the setting without a definition, and for cost reasons, we did not conduct the Lorem Ipsum definition experiment for GPT-4.} Within each of the two settings (binary/probabilistic), the best performing GPT-4 model is statistically significantly better than the other models.
GPT-4 performs equally well in the binary and the probabilistic setting (no statistically significant difference), with disagreement on only  7 out of 345 texts from the test set. Also, there is no significant difference compared to \texttt{TelConGBERT}. 

In contrast to GPT-4 and our expectations, GPT-3.5 does not achieve its best performance with a custom definition. It  attains its highest F1 score for the positive class in probabilistic prediction with the Lorem Ipsum definition.  While most of the performance differences for GPT-3.5 are not significant, e.g. providing no definition vs. Lorem Ipsum in the probabilistic setting, some are, e.g. probabilistic vs. binary setting using Lorem Ipsum. 
 
Llama 2 underperforms compared to both GPT models. We assume this to be due to the model's low exposure to non-English training data \cite{touvron_llama_2023_1}.
The model achieves its best F1 score on the positive class  without a definition, both in the binary and probabilistic settings. It produces similar scores for each definition in the probabilistic setting, but its performance varies greatly in the binary setting, ranging from F1 scores for the positive class from 0.08 to 0.63.

Further experiments showed that even minor and semantically negligible modifications of the prompt, such as changing the notation or the order of labels, impacted the performance of both GPT-3.5 and Llama 2. Additionally, formatting Llama 2's output in a parsable format was more difficult than for GPT models. Further investigation into this issue is required, e.g. to determine whether this is a language-independent issue. Moreover, that fact for both models, the optimal definition setting depends on the prediction setting, suggests that both are less robust than GPT-4.

All models, except for Llama 2 in one experiment, have higher F1 scores for class 0 compared to class 1, mirroring the trends observed in supervised fine-tuning. This outcome is expected due to the predominance of negative examples and their overall easier detection \cite{li2020}.

In summary, concerning RQ3, we can conclude that GPT-4's performance in the zero-shot setting with a custom definition of conspiracy narratives is comparable to that of the best supervised fine-tuned model, \texttt{TelConGBERT}.

\subsection{Few-Shot Classification}
Table \ref{tab:results_few_shot} shows that in the few-shot setting, all experiments produced inferior results compared to the corresponding zero-shot setting. These findings resonate with the conclusions drawn in a recent study by \citet{chae_large_2023}.
The F1 score for the positive class hovers around 0.7, while nearly 0.8 are achieved in zero-shot settings. 
It could be assumed that the lower performance in the few-shot setting is due to the lack of a definition. However, the performance also falls below that of zero-shot prompts without definition.  
Notably, there is no statistically significant advantage of GPT-4 over its predecessor GPT-3.5 in this setting.
In contrast, Llama 2 shows instability in few-shot scenarios, with high standard deviations. The model's outputs were also difficult to control, resulting in unusable data for analysis. Regarding the fact that Llama 2 was trained mainly on English-language data, its instability may be caused by the larger amount of German input in the few-shot setting. 

Few-shot experiments took 8 hours for GPT-3.5, 24 hours for GPT-4, and 15 hours for Llama 2.

\begin{table}[]
\caption{Few-shot performance  targeting binary label prediction. The values represent the mean $\pm$ standard deviation from ten runs using distinct training sets.}
\small
\begin{tabular}{p{0.82cm}p{0.65cm}|p{1.41cm}|p{1.41cm}|p{1.41cm}}&  &   \multicolumn{3}{c}{Mean $\pm$ SD}    \\ \hline
 &   Class    & GPT-3.5 & GPT-4 & Llama 2\\
 \hline
Precision               & 0     & $0.88\pm 0.03$  & $ \textbf{0.93}\pm 0.02 $     & $0.64\pm 0.26$      \\
 & 1     & $\textbf{0.59}\pm 0.04$       & $ \textbf{0.59} \pm 0.05 $    & $0.34\pm 0.06$       \\ \hline
\multirow{2}{*}{Recall} & 0     & $\textbf{0.72} \pm 0.06$       & $0.68 \pm 0.08$    & $0.29\pm 0.23$     \\
 & 1     & $0.80\pm 0.07$       & $\textbf{0.89} \pm 0.05$     & $0.75\pm 0.22$      \\ \hline
F1                 & 0     & $\textbf{0.79}\pm 0.03$       & $0.78 \pm 0.05$     & $0.36\pm 0.22$      \\ 
 & 1     & $0.68\pm 0.02$       & $\textbf{0.7} \pm 0.03$    & $0.45\pm 0.07$      \\ 
  & macro     & $0.73\pm 0.02$       & $\textbf{0.74} \pm 0.03$    & $0.41\pm 0.1$      \\\hline
\multicolumn{2}{l}{Accuracy}      & $\textbf{0.75}\pm 0.03$       & $\textbf{0.75} \pm 0.04$     &  $0.43\pm 0.1$   
\end{tabular}
\label{tab:results_few_shot}
\end{table}

\subsection{Comparative analysis}
As mentioned in Section \ref{sec:dataset}, the dataset TelCovACT encompasses information  whether a text communicating CTs alludes to the narrative components actor, strategy, and goal. Expert annotators faced the most challenges when the narrative was fragmented in the sense that not all three of these components were simultaneously present \cite{steffen_etal_2023}. This raises the question of whether detection models encounter the same difficulties.  
Furthermore, there is a broader question regarding the overlap in the models' predictions, particularly between \texttt{TelConGBERT} and the best prompt-based model (GPT-4, binary, custom definition). 

When tested against positive samples, both \texttt{TelConGBERT} and  GPT-4 demonstrate enhanced performance when at least two of the three component are simultaneously present (82\% and 88\%  detected, respectively) compared to highly fragmented narratives in which only one component was present (61\% and 69\% detected, respectively). This supports the hypothesis that increased fragmentation of the conspiracy narrative challenges the model's detection capabilities. 

Moreover, the prediction  probabilities of \texttt{TelConGBERT} and the output scores of the best GPT-4 model in probabilistic mode correlate with the fragmentation score of a text in the positive class, defined as the number of missing narrative components (thus ranging between 0 and 2), as shown in Table \ref{tab:fragmentation_score}. (Note  that the scores are only meaningful per model.) For GPT-3.5, on the contrary, no clear trend is visible. 

\begin{table}[h!]
\caption{Mean probability and probabilistic output score, respectively, grouped by  fragmentation score of the test data.}
\centering
\small
\begin{tabular}{l|lll}
 \makecell{fragmentation \\score} & TelConGBERT & GPT-4 & GPT-3.5 \\
\hline
 0     & 0.9  & 0.78 & 0.76     \\
1     & 0.85  & 0.77 & 0.78  \\
2     & 0.81  & 0.63 & 0.71
\end{tabular}
\label{tab:fragmentation_score}
\end{table}

While \texttt{TelConGBERT} and GPT-4 achieve comparable performance, their predictions do not align too well. In fact, the models do not agree on 15\% of the test data. Fragmentation, however, seems not to explain the divergence in assessment. It would be interesting to explore the differences in more detail, by e.g. allowing the prompting models to produce explanations and analyzing these qualitatively.

\subsection{Application of \texttt{TelConGBERT}}
\label{sec:application_Telcongbert}
As posited initially, models adept at detecting texts that propagate CTs can be invaluable for entities monitoring communications on both mainstream and fringe platforms. To indicate some  insights that can be gained from utilizing such a model in practice, we applied \texttt{TelConGBERT} to a total of 2,358,751 messages that were posted between March 11, 2020, and December 19, 2021, on one of 215 public channels that heavily focused on mobilization against Corona measures in German-speaking regions. Details regarding the channel selection can be found in the Appendix. 

The model estimated that an average of 11.74\% of all messages circulated CTs, translating to over a quarter-million such messages. In fact, the average frequency of messages per channel communicating CTs stood higher at 13.3\%, as one of the extremely populous channels, boasting more than 100,000 messages during the examined time frame, had a mere 2.5 messages predicted by the model as part of the positive class.
Delving deeper into the 178 channels that dispersed a minimum of 500 messages during this period, the ones most rife with conspiracy-laden communication were: `freiAuf' (with 40\% out of 1,323 messages), `DanielPrinzOffiziell' (38.7\% from 3,084 messages), and `stefanmagnet' (36.1\% out of 714 messages). On narrowing our focus to channels dispatching over 1,000 messages, the `ATTILAHILDMANN' channel, associated with its notorious namesake conspiracy theorist and antisemite, ranks third with 34.1\% of respective posts. A cursory glance at the descriptions of these channels corroborates the model's evaluations. For instance, `freiAuf', shorthand for `Freiheitliche Aufklärung' (engl.: Liberty enlightenment), headlines its Facebook page with the claim, ``if you're not convinced, watch this video which explains that the virus is a cover for 5G.'' `DanielPrinzOffiziell' is operated by Daniel Prinz, who gained notoriety through his book `Wenn das die Menschheit wüsste...' (engl.: If only mankind knew that ...), and promises to provide insights on the background to politics, Corona, and Deep State.

\section{Summary, Discussion and Future Work}

We comprehensively evaluated fine-tuning and prompting based approaches to classify Telegram posts obtained without keyword filters regarding the presence of conspiracy theories. Several of our modelling approaches demonstrate performance close to that of existing models for keyword-constrained English-language corpora. 
It is noteworthy that detecting conspiracy theories in Telegram posts is challenging, even for expert annotators, as evidenced by a Cohen's kappa value of 0.7 on the dataset utilized \cite{steffen_etal_2023}.\footnote{\citet{solopova_etal_2021} report $\kappa=0.65$ as interrater agreement on message-level assignment of categories of harmful language in data from one Telegram channel of Donald Trump supporters.} We thus encourage data compilation strategies that mitigate keyword bias and better reflect real-world application scenarios, even when dealing with challenging tasks and datasets.

Our best supervised fine-tuning approach \texttt{TelConGBERT} presents a viable and dependable choice that will be made available for  researchers and NGOs in this field.

With regard to RQ2, our evaluation of  temporal transfer scenarios within Telegram offers practical insights into model adaptation, suggesting that  models like \texttt{TelConGBERT} can be applied in real-world scenarios with modest additional annotation and fine-tuning efforts. 
In collaboration with NGOs, we will conduct a transdisciplinary research project aimed at optimizing the real-world deployment of \texttt{TelConGBERT} to monitor CTs on Telegram. We will investigate strategies for efficiently acquiring samples to update training data, methods for effectively communicating overall error rates and individual probabilities to end users, and mechanisms for collecting and integrating user feedback. 
These efforts address the current gap in practical applications of detection models for political texts (cf., e.g., \cite{salminen_etal_2021,kotarcic_etal_2023}), while exploring opportunities for transdisciplinary collaborations to maintain and improve these technologies. 
Furthermore, this work will expand the dataset TelCovACT by posts on different topics and from other Telegram channels. 
Extending it further by texts from other platforms would be beneficial, as a larger, more diverse corpus would allow for the exploration of CT detection in German on a more realistic corpus.

Nevertheless, it is essential to acknowledge that continuous annotation by experts requires resources and time, while exposing annotators to mental stress. 
Zero-shot classification using GPT-4, and even  GPT-3.5, offers an alternative with competitive performance that does not require explicit training data. However, it comes with its own set of challenges, primarily associated with high computational and monetary costs at prediction time, and its proprietary character. The decision regarding which approach to adopt ultimately hinges on the specific use case and available resources. Practitioners contemplating the integration of such models into real-world scenarios must carefully evaluate their needs and constraints to determine the optimal path \cite{chae_large_2023}.

Our findings have demonstrated that few-shot learning consistently produces suboptimal outcomes when compared to zero-shot scenarios. Additionally, this approach necessitates more time, resources, and financial investment than zero-shot learning.
As other research has shown, the decline in performance within few-shot settings might stem from the fact that ``some examples may negatively impact performance when compared to using the prompt alone, potentially due to their increased length and complexity'' \cite{chae_large_2023}. 
While strategic sampling might mitigate these negative impact, this method might not be practically viable in real-world scenarios, as argued in Section \ref{sec:methods_few_shot}. 
Nevertheless, alternative strategies for selecting in-context examples warrant further exploration. For instance, investigating aspects such as the impact of total input length could shed light on whether reduced examples (i.e., shorter input length) yield better results than longer inputs \cite{chae_large_2023,zhang_active_2022}.
Additionally, our experiments employed an equal distribution of examples from positive and negative classes. However, a well-balanced set of examples does not have to  consistently enhance performance or reduce variance. Some experiments even suggest that the model might not require exposure to examples for all labels   \cite{zhang_active_2022}. 
Considering this, experimenting with different label balances, such as providing only positive examples, could offer an approach applicable to real-world scenarios with limited financial resources.

The outputs of Llama 2 experiments were challenging to control and at times hard to explain, especially in the few-shot setting. Furthermore, the stark predominance of English pre-training data of the model may have contributed to the model's struggle with processing German-language input. Against this background, fine-tuning German-specific Llama 2 models as well as the utilization of other open models for German texts would be a promising  area for future work, hopefully allowing for results comparable to TelConGBERT and GPT-4 at lower mental and monetary cost. 

Another prospective direction for future research, particularly in examining differences between prompt-based and supervised fine-tuning approaches, entails analyzing the reasonings generated by prompting models. We have already conducted some exploratory experiments to obtain respective output, and are planning to continue further in this direction.
Nevertheless, expectations regarding achieving very high F1 scores should be toned down however. The task of CT detection remains complex due to the complexity of the phenomenon, and often defies binary classification. Acknowledging this complexity is vital, particularly regarding real-world application. 

\bibliography{references}

\begin{thebibliography}{75}
\expandafter\ifx\csname natexlab\endcsname\relax\def\natexlab#1{#1}\fi

\bibitem[{Aïmeur et~al.(2023)Aïmeur, Amri, and Brassard}]{ameur_etal_2023}
Esma Aïmeur, Sabrine Amri, and Gilles Brassard. 2023.
\newblock \href {https://doi.org/10.1007/s13278-023-01028-5} {Fake news,
  disinformation and misinformation in social media: a review}.
\newblock \emph{Social Network Analysis and Mining}, 13(1).

\bibitem[{Bang et~al.(2023)Bang, Cahyawijaya, Lee, Dai, Su, Wilie, Lovenia, Ji,
  Yu, Chung, Do, Xu, and Fung}]{bang_multitask_2023}
Yejin Bang, Samuel Cahyawijaya, Nayeon Lee, Wenliang Dai, Dan Su, Bryan Wilie,
  Holy Lovenia, Ziwei Ji, Tiezheng Yu, Willy Chung, Quyet~V. Do, Yan Xu, and
  Pascale Fung. 2023.
\newblock \href {http://arxiv.org/abs/2302.04023} {A {Multitask},
  {Multilingual}, {Multimodal} {Evaluation} of {ChatGPT} on {Reasoning},
  {Hallucination}, and {Interactivity}}.
\newblock ArXiv:2302.04023.

\bibitem[{Beltagy et~al.(2019)Beltagy, Lo, and Cohan}]{beltagy2019}
Iz~Beltagy, Kyle Lo, and Arman Cohan. 2019.
\newblock \href {https://doi.org/10.18653/v1/D19-1371} {{S}ci{BERT}: A
  pretrained language model for scientific text}.
\newblock In \emph{Proceedings of the 2019 Conference on Empirical Methods in
  Natural Language Processing and the 9th International Joint Conference on
  Natural Language Processing (EMNLP-IJCNLP)}, pages 3615--3620, Hong Kong,
  China. Association for Computational Linguistics.

\bibitem[{Bischoff et~al.(2022)Bischoff, Pustet, and
  Mihaljevi{\'c}}]{bischoff2022}
Nyco Bischoff, Milena Pustet, and Helena Mihaljevi{\'c}. 2022.
\newblock \href {https://doi.org/10.5281/ZENODO.6412114} {Datasheet for the
  dataset "{{Digitaler Hass}} - {{Antisemitismus}} und
  {{Verschw\"orungstheorien}} im {{Kontext}} der {{COVID-19 Pandemie}}"}.

\bibitem[{Chae and Davidson(2023)}]{chae_large_2023}
Youngjin Chae and Thomas Davidson. 2023.
\newblock \href {https://doi.org/10.31235/osf.io/sthwk} {Large {Language}
  {Models} for {Text} {Classification}: {From} {Zero}-{Shot} {Learning} to
  {Fine}-{Tuning}}.
\newblock Preprint, SocArXiv.

\bibitem[{Chen and Shu(2023)}]{chen_combating_2023}
Canyu Chen and Kai Shu. 2023.
\newblock \href {http://arxiv.org/abs/2311.05656} {Combating {Misinformation}
  in the {Age} of {LLMs}: {Opportunities} and {Challenges}}.
\newblock ArXiv:2311.05656.

\bibitem[{Chiu et~al.(2021)Chiu, Collins, and Alexander}]{chiu_detecting_2022}
Ke-Li Chiu, Annie Collins, and Rohan Alexander. 2021.
\newblock \href {http://arxiv.org/abs/2103.12407} {Detecting {Hate} {Speech}
  with {GPT}-3}.
\newblock ArXiv:2103.12407.

\bibitem[{Curley et~al.(2022)Curley, Siapera, and Carthy}]{curley2022a}
Cliona Curley, Eugenia Siapera, and Joe Carthy. 2022.
\newblock \href {https://doi.org/10.1177/20563051221129187} {Covid-19
  {{Protesters}} and the {{Far Right}} on {{Telegram}}: {{Co-Conspirators}} or
  {{Accidental Bedfellows}}?}
\newblock \emph{Social Media + Society}, 8(4):20563051221129187.

\bibitem[{Devlin et~al.(2019)Devlin, Chang, Lee, and Toutanova}]{devlin2018}
Jacob Devlin, Ming-Wei Chang, Kenton Lee, and Kristina Toutanova. 2019.
\newblock \href {https://doi.org/10.18653/v1/N19-1423} {{BERT}: Pre-training of
  deep bidirectional transformers for language understanding}.
\newblock In \emph{Proceedings of the 2019 Conference of the North {A}merican
  Chapter of the Association for Computational Linguistics: Human Language
  Technologies, Volume 1 (Long and Short Papers)}, pages 4171--4186,
  Minneapolis, Minnesota. Association for Computational Linguistics.

\bibitem[{Douglas et~al.(2017)Douglas, Sutton, and Cichocka}]{douglas2017}
Karen~M. Douglas, Robbie~M. Sutton, and Aleksandra Cichocka. 2017.
\newblock \href {https://doi.org/10.1177/0963721417718261} {The {{Psychology}}
  of {{Conspiracy Theories}}}.
\newblock \emph{Current Directions in Psychological Science}, 26(6):538--542.

\bibitem[{ElSherief et~al.(2021)ElSherief, Ziems, Muchlinski, Anupindi,
  Seybolt, De~Choudhury, and Yang}]{elsherief-etal-2021-latent}
Mai ElSherief, Caleb Ziems, David Muchlinski, Vaishnavi Anupindi, Jordyn
  Seybolt, Munmun De~Choudhury, and Diyi Yang. 2021.
\newblock \href {https://doi.org/10.18653/v1/2021.emnlp-main.29} {Latent
  hatred: A benchmark for understanding implicit hate speech}.
\newblock In \emph{Proceedings of the 2021 Conference on Empirical Methods in
  Natural Language Processing}, pages 345--363, Online and Punta Cana,
  Dominican Republic. Association for Computational Linguistics.

\bibitem[{Ernst et~al.(2017)Ernst, Engesser, B{\"u}chel, Blassnig, and
  Esser}]{ernst2017}
Nicole Ernst, Sven Engesser, Florin B{\"u}chel, Sina Blassnig, and Frank Esser.
  2017.
\newblock \href {https://doi.org/10.1080/1369118X.2017.1329333} {Extreme
  parties and populism: An analysis of {{Facebook}} and {{Twitter}} across six
  countries}.
\newblock \emph{Information, Communication \& Society}, 20:1--18.

\bibitem[{Gilardi et~al.(2023)Gilardi, Alizadeh, and
  Kubli}]{gilardi_chatgpt_2023}
Fabrizio Gilardi, Meysam Alizadeh, and Maël Kubli. 2023.
\newblock \href {http://arxiv.org/abs/2303.15056} {{ChatGPT} {Outperforms}
  {Crowd}-{Workers} for {Text}-{Annotation} {Tasks}}.
\newblock ArXiv:2303.15056.

\bibitem[{Gover et~al.(2020)Gover, Harper, and Langton}]{gover2020}
Angela~R. Gover, Shannon~B. Harper, and Lynn Langton. 2020.
\newblock \href {https://doi.org/10.1007/s12103-020-09545-1} {Anti-{{Asian Hate
  Crime During}} the {{COVID-19 Pandemic}}: {{Exploring}} the {{Reproduction}}
  of {{Inequality}}}.
\newblock \emph{American journal of criminal justice: AJCJ}, 45(4):647--667.

\bibitem[{Holzer(2021)}]{holzer2021}
Boris Holzer. 2021.
\newblock \href {https://doi.org/10.31235/osf.io/9rgtk} {{Zwischen Protest und
  Parodie: Strukturen der ``Querdenken''' -Kommunikation auf Telegram (und
  anderswo)}}.
\newblock In Sven Reichardt, editor, \emph{Die Misstrauensgemeinschaft der
  ``Querdenker'': Die Corona-Proteste aus kultur- und sozialwissenschaftlicher
  Perspektive}, pages 125--157. Campus Verlag, Frankfurt.

\bibitem[{Hoseini et~al.(2023)Hoseini, Melo, Benevenuto, Feldmann, and
  Zannettou}]{hoseini2021}
Mohamad Hoseini, Philipe Melo, Fabricio Benevenuto, Anja Feldmann, and Savvas
  Zannettou. 2023.
\newblock \href {https://doi.org/10.1145/3578503.3583603} {On the globalization
  of the qanon conspiracy theory through telegram}.
\newblock In \emph{Proceedings of the 15th ACM Web Science Conference 2023},
  WebSci '23, page 75–85, New York, NY, USA. Association for Computing
  Machinery.

\bibitem[{Huang et~al.(2023)Huang, Kwak, and An}]{huang_is_2023}
Fan Huang, Haewoon Kwak, and Jisun An. 2023.
\newblock \href {https://doi.org/10.1145/3543873.3587368} {Is {ChatGPT} better
  than {Human} {Annotators}? {Potential} and {Limitations} of {ChatGPT} in
  {Explaining} {Implicit} {Hate} {Speech}}.
\newblock In \emph{Companion {Proceedings} of the {ACM} {Web} {Conference}
  2023}, pages 294--297, Austin TX USA. ACM.

\bibitem[{{Idrissi-Yaghir} et~al.(2023){Idrissi-Yaghir}, Sch{\"a}fer, Bauer,
  and Friedrich}]{idrissi-yaghir2023}
Ahmad {Idrissi-Yaghir}, Henning Sch{\"a}fer, Nadja Bauer, and Christoph~M.
  Friedrich. 2023.
\newblock Domain {{Adaptation}} of {{Transformer-Based Models Using Unlabeled
  Data}} for {{Relevance}} and {{Polarity Classification}} of {{German Customer
  Feedback}}.
\newblock \emph{SN Computer Science}, 4(2):142.

\bibitem[{Japkowicz and Shah(2011)}]{japkowicz_shah_2011}
Nathalie Japkowicz and Mohak Shah. 2011.
\newblock \emph{Evaluating Learning Classifiers. A Classification Perspective}.
\newblock Cambridge University Press, New York, USA.

\bibitem[{Khatun and Brown(2023)}]{khatun2023reliability}
Aisha Khatun and Daniel~G. Brown. 2023.
\newblock \href {http://arxiv.org/abs/2306.06199} {Reliability check: An
  analysis of gpt-3's response to sensitive topics and prompt wording}.
\newblock ArXiv:2306.06199.

\bibitem[{Kotarcic et~al.(2023)Kotarcic, Hangartner, Gilardi, Kurer, and
  Donnay}]{kotarcic_etal_2023}
Ana Kotarcic, Dominik Hangartner, Fabrizio Gilardi, Selina Kurer, and Karsten
  Donnay. 2023.
\newblock \href {http://arxiv.org/abs/2212.02108} {Human-in-the-loop hate
  speech classification in a multilingual context}.
\newblock ArXiv: 2212.02108.

\bibitem[{Kou et~al.(2017)Kou, Gui, Chen, and Pine}]{kou2017}
Yubo Kou, Xinning Gui, Yunan Chen, and Kathleen Pine. 2017.
\newblock \href {https://doi.org/10.1145/3134696} {Conspiracy talk on social
  media: Collective sensemaking during a public health crisis}.
\newblock \emph{Proc. ACM Hum.-Comput. Interact.}, 1(CSCW).

\bibitem[{Kuznetsova et~al.(2023)Kuznetsova, Makhortykh, Vziatysheva, Stolze,
  Baghumyan, and Urman}]{kuznetsova2023generative}
Elizaveta Kuznetsova, Mykola Makhortykh, Victoria Vziatysheva, Martha Stolze,
  Ani Baghumyan, and Aleksandra Urman. 2023.
\newblock \href {http://arxiv.org/abs/2312.13096} {In generative ai we trust:
  Can chatbots effectively verify political information?}
\newblock ArXiv:2312.13096.

\bibitem[{Langguth et~al.(2023)Langguth, Schroeder, Filkuková, Brenner,
  Phillips, and Pogorelov}]{langguth_coco_2023}
Johannes Langguth, Daniel~Thilo Schroeder, Petra Filkuková, Stefan Brenner,
  Jesper Phillips, and Konstantin Pogorelov. 2023.
\newblock \href {https://doi.org/10.1007/s42001-023-00200-3} {{COCO}: an
  annotated {Twitter} dataset of {COVID}-19 conspiracy theories}.
\newblock \emph{Journal of Computational Social Science}, pages 1--42.

\bibitem[{Lee et~al.(2019)Lee, Yoon, Kim, Kim, Kim, So, and Kang}]{lee2019}
Jinhyuk Lee, Wonjin Yoon, Sungdong Kim, Donghyeon Kim, Sunkyu Kim, Chan~Ho So,
  and Jaewoo Kang. 2019.
\newblock \href {https://doi.org/10.1093/bioinformatics/btz682} {{{BioBERT}}: A
  pre-trained biomedical language representation model for biomedical text
  mining}.
\newblock \emph{Bioinformatics}, 36:1234–1240.

\bibitem[{Lei and Huang(2023)}]{lei2023identifying}
Yuanyuan Lei and Ruihong Huang. 2023.
\newblock \href {http://arxiv.org/abs/2310.18545} {Identifying conspiracy
  theories news based on event relation graph}.
\newblock ArXiv:2310.18545.

\bibitem[{Li et~al.(2023)Li, Fan, Atreja, and Hemphill}]{li_hot_2023}
Lingyao Li, Lizhou Fan, Shubham Atreja, and Libby Hemphill. 2023.
\newblock \href {http://arxiv.org/abs/2304.10619} {``{HOT}'' {ChatGPT}: {The}
  promise of {ChatGPT} in detecting and discriminating hateful, offensive, and
  toxic comments on social media}.
\newblock ArXiv:2304.10619.

\bibitem[{Li et~al.(2020)Li, Sun, Meng, Liang, Wu, and Li}]{li2020}
Xiaoya Li, Xiaofei Sun, Yuxian Meng, Junjun Liang, Fei Wu, and Jiwei Li. 2020.
\newblock \href {https://doi.org/10.18653/v1/2020.acl-main.45} {Dice loss for
  data-imbalanced {NLP} tasks}.
\newblock In \emph{Proceedings of the 58th Annual Meeting of the Association
  for Computational Linguistics}, pages 465--476, Online. Association for
  Computational Linguistics.

\bibitem[{Liu et~al.(2021)Liu, Shen, Zhang, Dolan, Carin, and
  Chen}]{liu_what_2021}
Jiachang Liu, Dinghan Shen, Yizhe Zhang, Bill Dolan, Lawrence Carin, and Weizhu
  Chen. 2021.
\newblock \href {http://arxiv.org/abs/2101.06804} {What {Makes} {Good}
  {In}-{Context} {Examples} for {GPT}-3?}
\newblock ArXiv:2101.06804.

\bibitem[{Liu et~al.(2023)Liu, Yuan, Fu, Jiang, Hayashi, and
  Neubig}]{liu_pre-train_2021}
Pengfei Liu, Weizhe Yuan, Jinlan Fu, Zhengbao Jiang, Hiroaki Hayashi, and
  Graham Neubig. 2023.
\newblock \href {https://doi.org/10.1145/3560815} {Pre-train, prompt, and
  predict: A systematic survey of prompting methods in natural language
  processing}.
\newblock \emph{ACM Comput. Surv.}, 55(9).

\bibitem[{Liu et~al.(2019)Liu, Ott, Goyal, Du, Joshi, Chen, Levy, Lewis,
  Zettlemoyer, and Stoyanov}]{liu2019}
Yinhan Liu, Myle Ott, Naman Goyal, Jingfei Du, Mandar Joshi, Danqi Chen, Omer
  Levy, Mike Lewis, Luke Zettlemoyer, and Veselin Stoyanov. 2019.
\newblock \href {http://arxiv.org/abs/1907.11692} {{{RoBERTa}}: {{A Robustly
  Optimized BERT Pretraining Approach}}}.
\newblock ArXiv:1907.11692.

\bibitem[{Liu et~al.(2024)Liu, Liu, Thompson, Yang, Jain, and
  Ananiadou}]{liu2024conspemollm}
Zhiwei Liu, Boyang Liu, Paul Thompson, Kailai Yang, Raghav Jain, and Sophia
  Ananiadou. 2024.
\newblock \href {http://arxiv.org/abs/2403.06765} {Conspemollm: Conspiracy
  theory detection using an emotion-based large language model}.
\newblock ArXiv:2403.06765.

\bibitem[{Mahl et~al.(2022)Mahl, Sch{\"a}fer, and Zeng}]{mahl2022a}
Daniela Mahl, Mike~S. Sch{\"a}fer, and Jing Zeng. 2022.
\newblock \href {https://doi.org/10.1177/14614448221075759} {Conspiracy
  theories in online environments: {{An}} interdisciplinary literature review
  and agenda for future research}.
\newblock \emph{New Media \& Society}, 25:1781--1801.

\bibitem[{Marcellino et~al.(2021)Marcellino, Helmus, Kerrigan, Reininger,
  Karimov, and Lawrence}]{marcellino2021}
William Marcellino, Todd~C. Helmus, Joshua Kerrigan, Hilary Reininger,
  Rouslan~I. Karimov, and Rebecca~Ann Lawrence. 2021.
\newblock \href {https://www.rand.org/pubs/research_reports/RRA676-1.html}
  {Detecting {{Conspiracy Theories}} on {{Social Media}}: {{Improving Machine
  Learning}} to {{Detect}} and {{Understand Online Conspiracy Theories}}}.
\newblock Technical report, {RAND Corporation}.

\bibitem[{Medina~Serrano et~al.(2020)Medina~Serrano, Papakyriakopoulos, and
  Hegelich}]{serrano2020}
Juan~Carlos Medina~Serrano, Orestis Papakyriakopoulos, and Simon Hegelich.
  2020.
\newblock \href {https://aclanthology.org/2020.nlpcovid19-acl.17} {{NLP}-based
  feature extraction for the detection of {COVID}-19 misinformation videos on
  {Y}ou{T}ube}.
\newblock In \emph{Proceedings of the 1st Workshop on {NLP} for {COVID-19} at
  {ACL} 2020}, Online. Association for Computational Linguistics.

\bibitem[{Memon and Carley(2020)}]{memon2020}
Shahan~Ali Memon and Kathleen~M. Carley. 2020.
\newblock \href {https://ceur-ws.org/Vol-2699/paper40.pdf} {Characterizing
  {COVID-19} misinformation communities using a novel twitter dataset}.
\newblock In \emph{Proceedings of the {CIKM} 2020 Workshops co-located with
  29th {ACM} International Conference on Information and Knowledge Management
  {(CIKM} 2020), Galway, Ireland, October 19-23, 2020}, volume 2699 of
  \emph{{CEUR} Workshop Proceedings}. CEUR-WS.org.

\bibitem[{Mendelsohn et~al.(2023)Mendelsohn, Le~Bras, Choi, and
  Sap}]{mendelsohn_dogwhistles_2023}
Julia Mendelsohn, Ronan Le~Bras, Yejin Choi, and Maarten Sap. 2023.
\newblock \href {https://doi.org/10.18653/v1/2023.acl-long.845} {From
  dogwhistles to bullhorns: Unveiling coded rhetoric with language models}.
\newblock In \emph{Proceedings of the 61st Annual Meeting of the Association
  for Computational Linguistics (Volume 1: Long Papers)}, pages 15162--15180,
  Toronto, Canada. Association for Computational Linguistics.

\bibitem[{Miani et~al.(2021)Miani, Hills, and Bangerter}]{miani_etal_2021}
Alessandro Miani, Thomas Hills, and Adrian Bangerter. 2021.
\newblock \href {https://doi.org/10.3758/s13428-021-01698-z} {Loco: The
  88-million-word language of conspiracy corpus}.
\newblock \emph{Behavior Research Methods}, 54(4):1794–1817.

\bibitem[{Moffitt et~al.(2021)Moffitt, King, and
  Carley}]{moffittHuntingConspiracyTheories2021}
J.~D. Moffitt, Catherine King, and Kathleen~M. Carley. 2021.
\newblock \href {https://doi.org/10.1177/20563051211043212} {Hunting
  {{Conspiracy Theories During}} the {{COVID-19 Pandemic}}}.
\newblock \emph{Social Media + Society}, 7(3).

\bibitem[{Mompelat et~al.(2022)Mompelat, Tian, Kessler, Luettgen, Rajanala,
  K{\"u}bler, and Seelig}]{mompelat2022loco}
Ludovic Mompelat, Zuoyu Tian, Amanda Kessler, Matthew Luettgen, Aaryana
  Rajanala, Sandra K{\"u}bler, and Michelle Seelig. 2022.
\newblock \href {https://aclanthology.org/2022.law-1.14} {How {``}loco{''} is
  the {LOCO} corpus? annotating the language of conspiracy theories}.
\newblock In \emph{Proceedings of the 16th Linguistic Annotation Workshop
  (LAW-XVI) within LREC2022}, pages 111--119, Marseille, France. European
  Language Resources Association.

\bibitem[{Mu et~al.(2023)Mu, Wu, Thorne, Robinson, Aletras, Scarton, Bontcheva,
  and Song}]{mu_navigating_2023}
Yida Mu, Ben~P. Wu, William Thorne, Ambrose Robinson, Nikolaos Aletras,
  Carolina Scarton, Kalina Bontcheva, and Xingyi Song. 2023.
\newblock \href {http://arxiv.org/abs/2305.14310} {Navigating {Prompt}
  {Complexity} for {Zero}-{Shot} {Classification}: {A} {Study} of {Large}
  {Language} {Models} in {Computational} {Social} {Science}}.
\newblock ArXiv: 2305.14310.

\bibitem[{Müller et~al.(2023)Müller, Salathé, and Kummervold}]{muller2020}
Martin Müller, Marcel Salathé, and Per~E. Kummervold. 2023.
\newblock \href {https://doi.org/10.3389/frai.2023.1023281}
  {{COVID-Twitter-BERT}: A natural language processing model to analyse
  {COVID-19} content on {Twitter}}.
\newblock \emph{Frontiers in Artificial Intelligence}, 6.

\bibitem[{Nguyen et~al.(2020)Nguyen, Vu, and Tuan~Nguyen}]{nguyen2020}
Dat~Quoc Nguyen, Thanh Vu, and Anh Tuan~Nguyen. 2020.
\newblock \href {https://doi.org/10.18653/v1/2020.emnlp-demos.2} {{{BERTweet}}:
  {{A}} pre-trained language model for {{English Tweets}}}.
\newblock In \emph{Proceedings of the 2020 {{Conference}} on {{Empirical
  Methods}} in {{Natural Language Processing}}: {{System Demonstrations}}},
  pages 9--14. {Association for Computational Linguistics}.

\bibitem[{Pan et~al.(2023)Pan, Wu, Lu, Luu, Wang, Kan, and
  Nakov}]{pan_fact-checking_2023}
Liangming Pan, Xiaobao Wu, Xinyuan Lu, Anh~Tuan Luu, William~Yang Wang, Min-Yen
  Kan, and Preslav Nakov. 2023.
\newblock \href {https://doi.org/10.18653/v1/2023.acl-long.386}
  {Fact-{Checking} {Complex} {Claims} with {Program}-{Guided} {Reasoning}}.
\newblock In \emph{Proceedings of the 61st {Annual} {Meeting} of the
  {Association} for {Computational} {Linguistics} ({Volume} 1: {Long}
  {Papers})}, pages 6981--7004, Toronto, Canada. Association for Computational
  Linguistics.

\bibitem[{Pelrine et~al.(2023)Pelrine, Imouza, Thibault, Reksoprodjo, Gupta,
  Christoph, Godbout, and Rabbany}]{pelrine2023reliable}
Kellin Pelrine, Anne Imouza, Camille Thibault, Meilina Reksoprodjo, Caleb
  Gupta, Joel Christoph, Jean-François Godbout, and Reihaneh Rabbany. 2023.
\newblock \href {http://arxiv.org/abs/2305.14928} {Towards reliable
  misinformation mitigation: Generalization, uncertainty, and gpt-4}.
\newblock ArXiv:2305.14928.

\bibitem[{Peskine et~al.(2023)Peskine, Papotti, and Troncy}]{peskine2021}
Youri Peskine, Paolo Papotti, and Rapha{\"e}l Troncy. 2023.
\newblock \href {https://ceur-ws.org/Vol-3181/paper65.pdf} {{Detection of
  COVID-19-Related Conpiracy Theories in Tweets using Transformer-Based Models
  and Node Embedding Techniques}}.
\newblock In \emph{{MediaEval 2022, Multimedia Evaluation Workshop, 12-13
  January 2023, Bergen, Norway}}, Bergen, Norway.

\bibitem[{Phillips et~al.(2022)Phillips, Ng, and Carley}]{phillips_etal_2022}
Samantha~C. Phillips, Lynnette Hui~Xian Ng, and Kathleen~M. Carley. 2022.
\newblock \href {https://doi.org/10.1145/3487553.3524665} {Hoaxes and hidden
  agendas: A twitter conspiracy theory dataset: Data paper}.
\newblock In \emph{Companion Proceedings of the Web Conference 2022}, WWW '22,
  page 876–880, New York, NY, USA. Association for Computing Machinery.

\bibitem[{Poddar et~al.(2024)Poddar, Mukherjee, Khatuya, Ganguly, and
  Ghosh}]{poddar2024covid19}
Soham Poddar, Rajdeep Mukherjee, Subhendu Khatuya, Niloy Ganguly, and Saptarshi
  Ghosh. 2024.
\newblock \href {http://arxiv.org/abs/2404.01669} {How {COVID-19} has impacted
  the anti-vaccine discourse: A large-scale {Twitter} study spanning
  pre-{COVID} and post-{COVID} era}.
\newblock ArXiv:2404.01669.

\bibitem[{Pogorelov et~al.(2021{\natexlab{a}})Pogorelov, Schroeder, Brenner,
  and Langguth}]{pogorelov2021a}
Konstantin Pogorelov, Daniel~Thilo Schroeder, Stefan Brenner, and Johannes
  Langguth. 2021{\natexlab{a}}.
\newblock \href {https://ceur-ws.org/Vol-3181/paper56.pdf} {{{FakeNews}}:
  {{Corona Virus}} and {{Conspiracies Multimedia Analysis Task}} at
  {{MediaEval}} 2021}.
\newblock In \emph{{MediaEval 2022, Multimedia Evaluation Workshop, 12-13
  January 2023, Bergen, Norway}}, Bergen, Norway.

\bibitem[{Pogorelov et~al.(2021{\natexlab{b}})Pogorelov, Schroeder,
  Filkukov\'{a}, Brenner, and Langguth}]{pogorelov_etal_2021}
Konstantin Pogorelov, Daniel~Thilo Schroeder, Petra Filkukov\'{a}, Stefan
  Brenner, and Johannes Langguth. 2021{\natexlab{b}}.
\newblock \href {https://doi.org/10.1145/3472720.3483617} {{WICO} text: A
  labeled dataset of conspiracy theory and 5g-corona misinformation tweets}.
\newblock In \emph{Proceedings of the 2021 Workshop on Open Challenges in
  Online Social Networks}, OASIS '21, page 21–25, New York, NY, USA.
  Association for Computing Machinery.

\bibitem[{Pustet and Mihaljević(2024)}]{pustet_etal_decoding_report_6_2024}
Milena Pustet and Helena Mihaljević. 2024.
\newblock Automated detection of antisemitic texts: is context all we need?
\newblock In \emph{Decoding Antisemitism: An AI-driven Study on Hate Speech and
  Imagery Online. Discourse Report 6}. Technical University Berlin. Centre for
  Research on Antisemitism.

\bibitem[{Reiss(2023)}]{reiss_testing_2023}
Michael~V. Reiss. 2023.
\newblock \href {http://arxiv.org/abs/2304.11085} {Testing the {Reliability} of
  {ChatGPT} for {Text} {Annotation} and {Classification}: {A} {Cautionary}
  {Remark}}.
\newblock ArXiv:2304.11085.

\bibitem[{Rivers and Lewis(2014)}]{rivers_ethical_2014}
Caitlin~M. Rivers and Bryan~L. Lewis. 2014.
\newblock \href {https://doi.org/10.12688/f1000research.3-38.v2} {Ethical
  research standards in a world of big data}.
\newblock Technical Report 3:38, F1000Research.

\bibitem[{Russo et~al.(2023)Russo, Stoehr, and Ribeiro}]{russo2023acti}
Giuseppe Russo, Niklas Stoehr, and Manoel~Horta Ribeiro. 2023.
\newblock \href {http://arxiv.org/abs/2307.06954} {Acti at evalita 2023:
  Overview of the conspiracy theory identification task}.
\newblock ArXiv:2307.06954.

\bibitem[{Sadler(2021)}]{sadler2021}
Neil Sadler. 2021.
\newblock \emph{Fragmented Narrative: Telling and Interpreting Stories in the
  {{Twitter}} Age}.
\newblock Critical Perspectives on Citizen Media. {Routledge}, {London; New
  York}.

\bibitem[{Salheiser and Richter(2020)}]{salheiser2020}
Axel Salheiser and Christoph Richter. 2020.
\newblock \href
  {https://www.idz-jena.de/fileadmin/user_upload/Factsheets/Factsheet_Proteste_Corona_Gefahr_Demokratie_Institut_f%C3%BCr_Demokratie_und_Zivilgesellschaft_Forschungsinstitut_Gesellschaftlicher_Zusammenhalt.pdf}
  {Factsheet: {{Poteste}} in der {{Corona-Pandemie}}: {{Gefahr}} f\"ur unsere
  {{Demokratie}}?}

\bibitem[{Salminen et~al.(2021)Salminen, Linarez, Jung, and
  Jansen}]{salminen_etal_2021}
Joni Salminen, Maria~Jose Linarez, Soon-gyo Jung, and Bernard~J. Jansen. 2021.
\newblock \href {https://doi.org/10.1109/BESC53957.2021.9635436} {Online hate
  detection systems: Challenges and action points for developers, data
  scientists, and researchers}.
\newblock In \emph{2021 8th International Conference on Behavioral and Social
  Computing (BESC)}, pages 1--7.

\bibitem[{Samory and Mitra(2018)}]{samory2018a}
Mattia Samory and Tanushree Mitra. 2018.
\newblock \href {https://doi.org/10.1145/3274421} {'{{The Government Spies
  Using Our Webcams}}': {{The Language}} of {{Conspiracy Theories}} in {{Online
  Discussions}}}.
\newblock \emph{Proceedings of the ACM on Human-Computer Interaction},
  2(CSCW):152:1--152:24.

\bibitem[{Shahsavari et~al.(2020)Shahsavari, Holur, Wang, Tangherlini, and
  Roychowdhury}]{shahsavari2020}
Shadi Shahsavari, Pavan Holur, Tianyi Wang, Timothy~R. Tangherlini, and Vwani
  Roychowdhury. 2020.
\newblock \href {https://doi.org/10.1007/s42001-020-00086-5} {Conspiracy in the
  time of corona: Automatic detection of emerging {{COVID-19}} conspiracy
  theories in social media and the news}.
\newblock \emph{Journal of Computational Social Science}, 3(2):279--317.

\bibitem[{Solopova et~al.(2021)Solopova, Scheffler, and
  Popa-Wyatt}]{solopova_etal_2021}
Veronika Solopova, Tatjana Scheffler, and Mihaela Popa-Wyatt. 2021.
\newblock \href {https://doi.org/10.5334/johd.32} {A telegram corpus for hate
  speech, offensive language, and online harm}.
\newblock \emph{Journal of Open Humanities Data}, 7.

\bibitem[{Steffen et~al.(2023)Steffen, Mihaljevic, Pustet, Castro~Varela,
  Bischoff, Bayramoglu, and Oghalai}]{steffen_etal_2023}
Elisabeth Steffen, Helena Mihaljevic, Milena Pustet, Maria do~Mar
  Castro~Varela, Nyco Bischoff, Yener Bayramoglu, and Bahar Oghalai. 2023.
\newblock \href
  {https://https://www.icwsm.org/2023/index.html/accepted_papers.html} {Codes,
  patterns and shapes of contemporary online antisemitism and conspiracy
  narratives. an annotation guide and labeled german-language dataset in the
  context of covid-19}.
\newblock In \emph{Proceedings of the 23rd International AAAI Conference on Web
  and Social Media}, pages 1--11, Cyprus.

\bibitem[{Sunstein and Vermeule(2009)}]{sunstein2009}
Cass~R. Sunstein and Adrian Vermeule. 2009.
\newblock \href {https://doi.org/10.1111/j.1467-9760.2008.00325.x} {Conspiracy
  {{Theories}}: {{Causes}} and {{Cures}}*}.
\newblock \emph{Journal of Political Philosophy}, 17(2):202--227.

\bibitem[{Tian et~al.(2023)Tian, Mitchell, Zhou, Sharma, Rafailov, Yao, Finn,
  and Manning}]{tian2023just}
Katherine Tian, Eric Mitchell, Allan Zhou, Archit Sharma, Rafael Rafailov,
  Huaxiu Yao, Chelsea Finn, and Christopher Manning. 2023.
\newblock \href {https://doi.org/10.18653/v1/2023.emnlp-main.330} {Just ask for
  calibration: Strategies for eliciting calibrated confidence scores from
  language models fine-tuned with human feedback}.
\newblock In \emph{Proceedings of the 2023 Conference on Empirical Methods in
  Natural Language Processing}, pages 5433--5442, Singapore. Association for
  Computational Linguistics.

\bibitem[{Touvron et~al.(2023)Touvron, Martin, Stone, Albert, Almahairi,
  Babaei, Bashlykov, Batra, Bhargava, Bhosale, Bikel, Blecher, Ferrer, Chen,
  Cucurull, Esiobu, Fernandes, Fu, Fu, Fuller, Gao, Goswami, Goyal, Hartshorn,
  Hosseini, Hou, Inan, Kardas, Kerkez, Khabsa, Kloumann, Korenev, Koura,
  Lachaux, Lavril, Lee, Liskovich, Lu, Mao, Martinet, Mihaylov, Mishra,
  Molybog, Nie, Poulton, Reizenstein, Rungta, Saladi, Schelten, Silva, Smith,
  Subramanian, Tan, Tang, Taylor, Williams, Kuan, Xu, Yan, Zarov, Zhang, Fan,
  Kambadur, Narang, Rodriguez, Stojnic, Edunov, and
  Scialom}]{touvron_llama_2023_1}
Hugo Touvron, Louis Martin, Kevin Stone, Peter Albert, Amjad Almahairi, Yasmine
  Babaei, Nikolay Bashlykov, Soumya Batra, Prajjwal Bhargava, Shruti Bhosale,
  Dan Bikel, Lukas Blecher, Cristian~Canton Ferrer, Moya Chen, Guillem
  Cucurull, David Esiobu, Jude Fernandes, Jeremy Fu, Wenyin Fu, Brian Fuller,
  Cynthia Gao, Vedanuj Goswami, Naman Goyal, Anthony Hartshorn, Saghar
  Hosseini, Rui Hou, Hakan Inan, Marcin Kardas, Viktor Kerkez, Madian Khabsa,
  Isabel Kloumann, Artem Korenev, Punit~Singh Koura, Marie-Anne Lachaux,
  Thibaut Lavril, Jenya Lee, Diana Liskovich, Yinghai Lu, Yuning Mao, Xavier
  Martinet, Todor Mihaylov, Pushkar Mishra, Igor Molybog, Yixin Nie, Andrew
  Poulton, Jeremy Reizenstein, Rashi Rungta, Kalyan Saladi, Alan Schelten, Ruan
  Silva, Eric~Michael Smith, Ranjan Subramanian, Xiaoqing~Ellen Tan, Binh Tang,
  Ross Taylor, Adina Williams, Jian~Xiang Kuan, Puxin Xu, Zheng Yan, Iliyan
  Zarov, Yuchen Zhang, Angela Fan, Melanie Kambadur, Sharan Narang, Aurelien
  Rodriguez, Robert Stojnic, Sergey Edunov, and Thomas Scialom. 2023.
\newblock \href {http://arxiv.org/abs/2307.09288} {Llama 2: {Open} {Foundation}
  and {Fine}-{Tuned} {Chat} {Models}}.
\newblock ArXiv:2307.09288.

\bibitem[{Tunstall et~al.(2022)Tunstall, von Werra, Wolf, and
  G{\'e}ron}]{tunstall2022}
Lewis Tunstall, Leandro von Werra, Thomas Wolf, and Aur{\'e}lien G{\'e}ron.
  2022.
\newblock \emph{Natural Language Processing with {{Transformers}}: Building
  Language Applications with {{Hugging Face}}}, first edition edition.
\newblock {O'Reilly}, {Beijing Boston Farnham Sebastopol Tokyo}.

\bibitem[{Törnberg(2023)}]{tornberg_chatgpt_2023}
Petter Törnberg. 2023.
\newblock \href {http://arxiv.org/abs/2304.06588} {{ChatGPT}-4 {Outperforms}
  {Experts} and {Crowd} {Workers} in {Annotating} {Political} {Twitter}
  {Messages} with {Zero}-{Shot} {Learning}}.
\newblock ArXiv:2304.06588.

\bibitem[{Vaigh et~al.(2021)Vaigh, Girault, Mallart, and Nguyen}]{vaigh2021}
Cheikh Brahim~El Vaigh, Thomas Girault, Cyrielle Mallart, and Duc~Hau Nguyen.
  2021.
\newblock \href {https://ceur-ws.org/Vol-3181/paper68.pdf} {Detecting {{Fake
  News Conspiracies with Multitask}} and {{Prompt-Based Learning}}}.
\newblock In \emph{Working Notes Proceedings of the MediaEval 2021 Workshop}.
  CEUR.

\bibitem[{Vergani et~al.(2022)Vergani, Martinez~Arranz, Scrivens, and
  Orellana}]{vergani2022}
Matteo Vergani, Alfonso Martinez~Arranz, Ryan Scrivens, and Liliana Orellana.
  2022.
\newblock \href {https://doi.org/10.1177/20563051221138758} {Hate {{Speech}} in
  a {{Telegram Conspiracy Channel During}} the {{First Year}} of the {{COVID-19
  Pandemic}}}.
\newblock \emph{Social Media + Society}, 8(4).

\bibitem[{Wang and Chang(2022)}]{wang_toxicity_2022}
Yau-Shian Wang and Yingshan Chang. 2022.
\newblock \href {http://arxiv.org/abs/2205.12390} {Toxicity {Detection} with
  {Generative} {Prompt}-based {Inference}}.
\newblock ArXiv:2205.12390.

\bibitem[{White et~al.(2023)White, Fu, Hays, Sandborn, Olea, Gilbert, Elnashar,
  Spencer-Smith, and Schmidt}]{white_prompt_2023}
Jules White, Quchen Fu, Sam Hays, Michael Sandborn, Carlos Olea, Henry Gilbert,
  Ashraf Elnashar, Jesse Spencer-Smith, and Douglas~C. Schmidt. 2023.
\newblock \href {http://arxiv.org/abs/2302.11382} {A {Prompt} {Pattern}
  {Catalog} to {Enhance} {Prompt} {Engineering} with {ChatGPT}}.
\newblock ArXiv:2302.11382.

\bibitem[{Winter et~al.(2021)Winter, Gerster, Helmer, and Baaken}]{winter2021}
Hannah Winter, Lea Gerster, Joschua Helmer, and Till Baaken. 2021.
\newblock \href
  {https://www.isdglobal.org/isd-publications/uberdosis-desinformation-die-vertrauenskrise-impfskepsis-und-impfgegnerschaft-in-der-covid-19-pandemie/}
  {{\"Uberdosis Desinformation: Die Vertrauenskrise. Impfskepsis und
  Impfgegnerschaft in der COVID-19-Pandemie}}.
\newblock Technical report, Institute for Strategic Dialogue.

\bibitem[{Yu et~al.(2023)Yu, Yang, Pelrine, Godbout, and
  Rabbany}]{yu_open_2023}
Hao Yu, Zachary Yang, Kellin Pelrine, Jean~Francois Godbout, and Reihaneh
  Rabbany. 2023.
\newblock \href {http://arxiv.org/abs/2308.10092} {Open, {Closed}, or {Small}
  {Language} {Models} for {Text} {Classification}?}
\newblock ArXiv:2308.10092.

\bibitem[{Zeitung(2021)}]{zeitungsueddeutsche2021}
S{\"u}ddeutsche Zeitung. 2021.
\newblock \href
  {https://www.sueddeutsche.de/politik/youtube-querdenker-kanal-geloescht-1.5304431}
  {{"Querdenker"-Kanal gel\"oscht}}.
\newblock \emph{S\"uddeutsche.de}.
\newblock May 26, 2021.

\bibitem[{Zhang et~al.(2022)Zhang, Feng, and Tan}]{zhang_active_2022}
Yiming Zhang, Shi Feng, and Chenhao Tan. 2022.
\newblock \href {https://doi.org/10.18653/v1/2022.emnlp-main.622} {Active
  {Example} {Selection} for {In}-{Context} {Learning}}.
\newblock In \emph{Proceedings of the 2022 {Conference} on {Empirical}
  {Methods} in {Natural} {Language} {Processing}}, pages 9134--9148, Abu Dhabi,
  United Arab Emirates. Association for Computational Linguistics.

\bibitem[{Zhou and Zafarani(2020)}]{zhou_etal_fake_news_2020}
Xinyi Zhou and Reza Zafarani. 2020.
\newblock \href {https://doi.org/10.1145/3395046} {A survey of fake news:
  Fundamental theories, detection methods, and opportunities}.
\newblock \emph{ACM Computing Surveys}, 53(5).

\end{thebibliography}

\appendix

\section{Appendix}

\subsection{Ethical Considerations}

Our research adheres to established ethical standards and is guided by best practices outlined in \cite{rivers_ethical_2014}. Our work is centered on enhancing methods for the detection of harmful content, ultimately contributing to the reduction of negative impacts associated with online communication.
The data employed in our experiments was thoroughly collected and processed, adhering to established best practices, namely gathering only publicly available data and ensuring that no information could be used to identify authors or individuals. Moreover, the data utilized for model training is available upon request and adheres to FAIR principles \cite{bischoff2022}.
In the context of our research, it is essential to acknowledge the inherent challenges associated with the deployment of AI models. Model errors can have negative consequences, especially when applied in real-world contexts: False positives may penalize counter speech or lead to unjustified regulations or sanctions on users. Conversely, detection algorithms are vulnerable to strategic deception by malicious actors, which might increase the number of false negatives and therefore proliferate the dissemination of CT content instead of mitigating it.

In contrast to NLU-oriented models like our best performing model \texttt{TelConGBERT}, the use of generative models in this context presents unique ethical considerations, as they can potentially be misused to produce harmful content. We emphasize that our experiments did not request models to generate such content, and that providers of these models have implemented guardrails to prevent misuse.

Further ethical challenges stem from the limited transparency of closed models, and the costs\footnote{The total cost of our experiments using models from OpenAI amounted to around 500 Dollars.} associated with their usage, resulting in severe limitations of accessibility for e.g. smaller monitoring NGOs who could benefit from automated detection methods, but only have limited resources.
To address these concerns, we will make our best model publicly available under a permissive license, to promote accessibility and usage among organizations with limited resources. 

\subsection{Fine-Tuning of Transformer Models}
\begin{sloppypar}
\paragraph{Initial Experiment} The following pre-trained models were assessed: \texttt{bert-base-multilingual-cased}, \texttt{bert-base-multilingual-uncased}, \texttt{deepset/gbert-base}, \texttt{deepset/gbert-large}, \texttt{distilbert-base-multilingual-cased}, \texttt{distilbert-base-german-cased}, \texttt{xlm-roberta-base}, \texttt{xlm-roberta-large},  \texttt{uklfr/gottbert-base}.
We used the following hyperparameter setting: both dropout probabilities set to 0.1, batch size of 16, learning rate set to 5e-05, no weight decay, trained for 8 epochs. 
For all models the validation loss starts to grow after 2 epochs latest. We thus evaluated the models in terms of the F1 score on class 1 and the macro F1 score based on the first 2 epochs. 
\end{sloppypar}

\paragraph{Hyperparameter Optimization} 
A Bayesian optimization of the best performing pre-trained model was employed to assess various combinations of model and dataset-related hyperparameters. 
 Specifically, we examined the impact of emojis and channel-specific footers; we created a balanced variant of the training data by randomly downsampling the negative class; and we allowed adjustments of typical  model-specific hyperparameters. Fine-tuning was limited to a maximum of 4 epochs as fine-tuning for classification tasks on small datasets typically converges after 2 to 3 epochs. The optimization procedure encompassed 600 iterations, aiming to minimize the cross-entropy loss on the validation set. 
We additionally conducted a grid search within a narrowed hyperparameter space informed by the results of the Bayesian optimization to assess the tradeoff between computing time efficiency and performance improvement.
Moreover, Bayesian hyperparameter tuning was repeated with the self-adjusting dice loss \cite{li2020} which should be more immune to the data-imbalance issue than cross-entropy loss. The parameter $\alpha$ that regulates the weight of easy examples during training was in the range between 0 and 0.7.

All experiments were run on a server equipped with two Nvidia A30 GPUs, an Intel(R) Xeon(R) Gold 6346 CPU, and 251 GB RAM. Details concerning fine-tuning can be found in the Appendix.

The grid search ran for 12 days on a single Nvidia A30 GPU to complete almost 7,000 runs, while the Bayesian optimization with 600 runs completed within 1 day. Since the latter yielded a model with measured scores lowered only by 0.01, this would be the recommended approach in practice.

\paragraph{Model Retraining}
\label{subsec:model_retraining}

We utilized the corpus from which the annotated TelCovACT dataset was crafted \cite{steffen_etal_2023}, encompassing $\sim\! 1.35$ million messages from 215 public Telegram channels. The records were pre-processed by removing URLs, user handles, IBANs, and trailing white spaces as well as duplicate texts and those with less than five tokens. The remaining data was split at an 8:1 ratio into a training (1,199,643 records) and a validation (149,956 records) set.  
The best performing model with regard to the initial experiment was further pre-trained over 20 epochs on the Masked Language Model (MLM) task only, enabling a shorter training time without a negative impact on downstream tasks  \cite{idrissi-yaghir2023,liu2019,tunstall2022}.
The tokenizer vocabulary was left unmodified, since the addition of in-domain vocabulary, if it is not expected to differ substantially, has a rather limited impact  \cite{beltagy2019,idrissi-yaghir2023}. To achieve faster training, the maximal sequence length of the inputs was reduced to 128 as this fits well the length of typical messages in our corpus. 
The learning rate was set to 2e-5 as proposed by \cite{muller2020}, and the remaining hyperparameters were left at their default values.
The retraining encompassed 20 epochs and took approximately 3.5 days on an Nvidia A30 GPU, with validation loss decreasing from 1.71 to 1.46.

\subsection{Zero-Shot and Few-Shot Experiments}\label{appendix:def_prompts}

\paragraph{Conspiracy Theory Definition}
Conspiracy theories formulate the strong belief that a secret group of people, who have the evil goal of taking over institutions, countries, or the world, intentionally cause complex, and in most cases unsolved, events and phenomena. 
Conspiracy theories can be considered an effort to explain some event or practice by reference to the machinations of powerful people, 
who have managed to conceal their role. Such a narrative is based on a simple dualism between good and evil which leaves no space for unintentional, 
unforeseeable things or mistakes to happen. A conspiracy theory typically involves actors who use a strategy to pursue a concrete malicious goal. 
Often, conspiracy theories are communicated in a fragmented way, so that not all of these components need to be present in a text. 
In some cases, a conspiracy theory is not explicitly articulated, but only referenced in a text via certain codes or hashtags.

\paragraph{System Prompt} `You are a data annotation expert trained to identify conspiracy theories on social media.'

\paragraph{Hyperparameters}
temperature: 0 (GPT models) and 0.01 (Llama 2); 
footers removed and emojis kept for all models.

\begin{table}[h!]
\caption{Prompts for zero-shot binary classification. The bold part of the instruction is replaced by `or not' in those experiments, where no definition is provided.}
\small
\begin{tabular}{|p{1.2cm}|p{2.5cm}|p{2.5cm}|}
\hline
\multicolumn{1}{p{1.2cm}|}{Model} &
\multicolumn{1}{p{2.5cm}|}{GPT-3.5 \& GPT-4} &
\multicolumn{1}{p{2.5cm}}{Llama 2} \\ \hline
\multicolumn{1}{p{1.2cm}|}{Instruction} &
\multicolumn{2}{p{5.8cm}}{Consider the following message: `\{message\}'. You have to decide whether the message communicates a conspiracy theory \textbf{considering the following definition: `\{definition\}'}. Give your answer using one of the two options: \newline a) Yes \newline b) No} \\ \hline
\multicolumn{1}{p{1.2cm}|}{Output constraint} &
\multicolumn{1}{p{2.5cm}|}{Do not provide any other outputs or any explanation for your output.} &
\multicolumn{1}{p{2.5cm}}{Answer in one line, only use Yes or No.} \\ \hline
\end{tabular}
\label{tab:prompts_binary}
\end{table}

\begin{table}[h]
\caption{Prompts for zero-shot probabilistic classification. The bold part of the instruction is replaced by `or not' in those experiments, where no definition is provided.}
\small
\begin{tabular}{|p{1.2cm}|p{2.5cm}|p{2.5cm}|}
\hline
\multicolumn{1}{p{1.2cm}|}{Model} &
\multicolumn{1}{p{2.5cm}|}{GPT-3.5 \& GPT-4} &
\multicolumn{1}{p{2.5cm}}{Llama 2} \\ \hline
\multicolumn{1}{p{1.2cm}|}{Instruction} &
  \multicolumn{2}{p{5.8cm}}{Consider the following message:'\{message\}'.
You have to decide whether the message communicates a conspiracy theory \textbf{(considering the following definition: '\{definition\}'}.
I want you to provide a probability score between 0 to 1 where the score represents the probability that the message communicates a conspiracy theory. A probability of 1 means that the comment is highly likely to communicate a conspiracy theory.} \\ \hline
\multicolumn{1}{p{1.2cm}|}{Output constraint} & \multicolumn{1}{p{2.5cm}|}{Do not provide any other outputs or any explanation for your output.} & \multicolumn{1}{p{2.5cm}}{Answer in one line, only return the score. Do not provide any other outputs or any explanation for your output. The score is: } \\ \hline
\end{tabular}
\label{tab:prompts_proba}
\end{table}

\begin{table}[h]
\caption{Prompts for few-shot binary classification.}
\small
\begin{tabular}{|p{1.2cm}|p{2.5m}|p{2.5cm}|}
\hline
\multicolumn{1}{p{1.2cm}|}{Model} &
\multicolumn{1}{p{2.5cm}|}{GPT-3.5 \& GPT-4} &
\multicolumn{1}{p{2.5cm}}{Llama 2} \\ \hline
\multicolumn{1}{p{1.2cm}|}{Instruction including few-shot examples} &
\multicolumn{2}{p{5.8cm}}{You have to decide whether the message communicates a conspiracy theory or not. \newline  
Examples: \newline
message: \{message\_1\} \newline
label: \{label\_1\} \newline
… \newline
message: \{message\_14\} \newline
label: \{label\_14\} } \\ \hline
\multicolumn{1}{p{1.2cm}|}{Output constraint} &
\multicolumn{1}{p{2.5cm}|}{message: \{message\} \newline label: } &
\multicolumn{1}{p{2.5cm}}{Answer in one line, only return the label. \newline message: \{message\} \newline Label: } \\ \hline
\end{tabular}
\label{tab:prompts_few}
\end{table}

\newpage
\subsection{Telegram Channels with Focus on Mobilization Against COVID-19 Measures}
In Section \ref{sec:application_Telcongbert}, we applied the model \texttt{TelConGBERT} to a corpus comprising 215 public Telegram channels. The selection of these channels is described in detail in the datasheet of the dataset TelCovACT \cite{bischoff2022} which we utilized for model training. The method for channel selection was roughly as follows: firstly, all channels identified as relevant for mobilization against Corona measures in a research report \cite{salheiser2020} during the pandemic's early phase that had a minimum of 1,000 followers were selected. Additionally, channels mentioned in tweets related to the ‘Querdenken’ movement against Corona measures from three distinctive periods centering around pivotal demonstrations in 2020 and 2021 were added. The dataset  TelCovACT itself was sampled from a subset of these channels. 
\end{document}